\documentclass[conference]{IEEEtran}
\IEEEoverridecommandlockouts
\usepackage{cite}
\usepackage{amsmath,amssymb,amsfonts}
\usepackage{algorithmic}
\usepackage{graphicx}
\usepackage{textcomp}
\usepackage[table]{xcolor}
\usepackage{booktabs} 
\usepackage{multirow}
\usepackage{pifont}
\newcommand{\cmark}{\texttt{\ding{51}}}

\def\BibTeX{{\rm B\kern-.05em{\sc i\kern-.025em b}\kern-.08em
    T\kern-.1667em\lower.7ex\hbox{E}\kern-.125emX}}
\begin{document}

\title{AU-TTT: Vision Test-Time Training model for Facial Action Unit Detection}

\author{
Bohao Xing\textsuperscript{1,*},
Kaishen Yuan\textsuperscript{2,*},
Zitong Yu\textsuperscript{3},
Xin Liu\textsuperscript{1,\dag},
Heikki Kälviäinen\textsuperscript{1,4,5} \\
\textsuperscript{1}Lappeenranta-Lahti University of Technology LUT, Finland \\
\textsuperscript{2}The Hong Kong University of Science and Technology (Guangzhou), China  \textsuperscript{3}Great Bay University, China \\
\textsuperscript{4}Rensselaer Polytechnic Institute, USA \textsuperscript{5}Brno University of Technology, Czech Republic \\
\texttt{\{bohao.xing, xin.liu, heikki.kalviainen\}@lut.fi,}\\
\texttt{yuankaishen01@gmail.com, zitong.yu@ieee.org} \\
\thanks{This work is supported in part by the China Scholarship Council (Grant No.~202406250043), and the Finnish IT Center for Science is gratefully acknowledged.}
\thanks{\textsuperscript{*}Equal contributions.}
\thanks{\textsuperscript{\dag}Corresponding author.}
}

\maketitle
\begin{abstract}
Facial Action Units (AUs) detection is a cornerstone of objective facial expression analysis and a critical focus in affective computing. Despite its importance, AU detection faces significant challenges, such as the high cost of AU annotation and the limited availability of datasets. These constraints often lead to overfitting in existing methods, resulting in substantial performance degradation when applied across diverse datasets. Addressing these issues is essential for improving the reliability and generalizability of AU detection methods.
Moreover, many current approaches leverage Transformers for their effectiveness in long-context modeling, but they are hindered by the quadratic complexity of self-attention. Recently, Test-Time Training (TTT) layers have emerged as a promising solution for long-sequence modeling. Additionally, TTT applies self-supervised learning for iterative updates during both training and inference, offering a potential pathway to mitigate the generalization challenges inherent in AU detection tasks.
In this paper, we propose a novel vision backbone tailored for AU detection, incorporating bidirectional TTT blocks, named AU-TTT. Our approach introduces TTT Linear to the AU detection task and optimizes image scanning mechanisms for enhanced performance. Additionally, we design an AU-specific Region of Interest (RoI) scanning mechanism to capture fine-grained facial features critical for AU detection. Experimental results demonstrate that our method achieves competitive performance in both within-domain and cross-domain scenarios.
\end{abstract}


\section{Introduction}
\label{sec:intro}

Facial action units, as defined by the Facial Action Coding System, represent specific movements or deformations of facial muscles, providing a framework for analyzing complex facial expressions~\cite{martinez2017automatic,xing2024emo,li2024enhancing,li2024counterfactual,li2024eald}. Consequently, AU detection has become a highly active area of research in recent years, spurring the development of numerous advanced methods. Traditionally, most approaches have relied on Convolutional Neural Networks (CNNs)\cite{he2016deep} or Graph Neural Networks (GNNs)\cite{scarselli2008graph}, which limit the model’s focus to localized facial regions. Recently, however, innovative methods~\cite{jacob2021facial, liu2024multi, luo2022learning, wang2022semantic, yang2023fan, yuan2025auformer, wang2024multi} have introduced Transformers~\cite{dosovitskiy2020image}, leveraging their exceptional capability to model long-range dependencies and extract richer global context information from facial data.

\begin{figure}[t]
    \centering
    \includegraphics[width=0.6\linewidth]{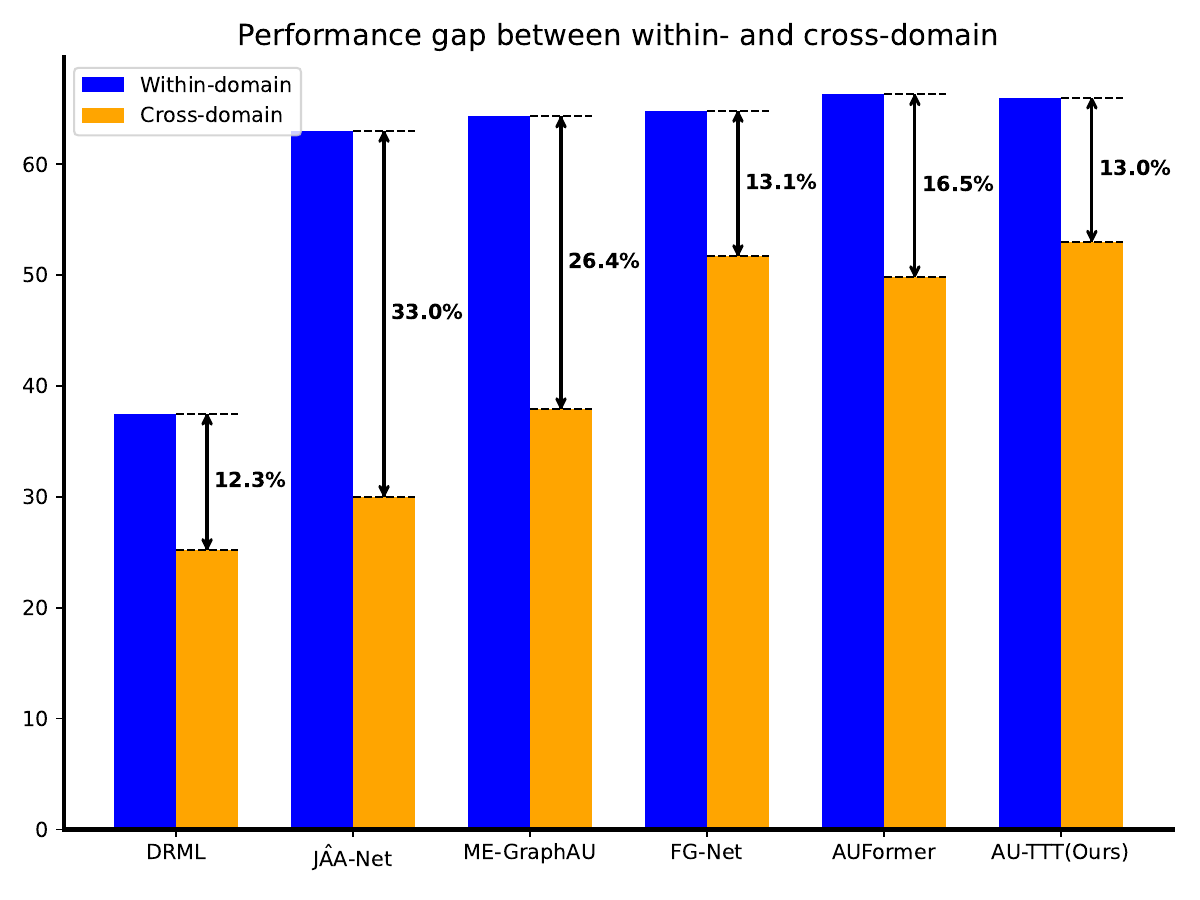} 
    \vspace{-1.5em}
    \caption{Performance (F1 score) gap between the within- and cross-domain AU detection for DRML~\cite{zhao2016deep}, J$\hat{\mathrm{A}}$A-Net~\cite{shao2021jaa}, MEGraphAU~\cite{luo2022learning}, FG-Net~\cite{yin2024fg}, AUFormer~\cite{yuan2025auformer}, and AU-TTT(Ours). The within-domain performance is averaged between DISFA and BP4D, while the cross-domain performance is averaged between BP4D to DISFA and DISFA to BP4D.}
    \vspace{-1.5em}
    \label{fig:performance_gap} 
\end{figure}

Specifically, early AU detection methods relied heavily on hand-crafted features, which struggled to capture the complexity and subtlety of facial muscle activations, resulting in limited performance. The advent of deep learning, particularly CNNs, marked a significant breakthrough by enabling automatic feature extraction. 
Methods such as J$\hat{\mathrm{A}}$A-Net~\cite{shao2021jaa} employed multi-scale hierarchical designs to capture features at varying resolutions, thereby improving the detection of AUs across diverse scales of muscle activation.

To address the intricate relationships among AUs, researchers explored GNNs\cite{luo2022learning, niu2019local}, which effectively model the inherent dependencies between AUs using graph-based structures. 
For example, ME-GraphAU\cite{luo2022learning} utilized a graph with multi-dimensional edge features and Gated Graph Convolutional Networks~\cite{bresson2017residual} to capture correlations between AU pairs, leading to enhanced detection accuracy. 
However, despite their effectiveness, GNN-based methods primarily emphasize local features and fail to fully exploit the global dependencies present in the data.

The recent integration of Transformers into AU detection has significantly enhanced the field by enabling the modeling of long-range dependencies~\cite{jacob2021facial, wang2022semantic, yang2023fan, liu2024multi, wang2024multi, yuan2025auformer}. 
For instance, methods like FAUDT~\cite{jacob2021facial} leveraged Transformer blocks to capture global correlations between AUs, thereby providing richer contextual information for more accurate detection.
However, despite these advancements, the substantial number of learnable parameters in Transformer-based models poses a challenge, increasing the risk of overfitting, particularly when training on datasets with limited AU annotations. Furthermore, the quadratic complexity of the self-attention mechanism significantly increases computational costs.

\begin{figure*}[t]
    \centering
    \includegraphics[width=0.5\linewidth]{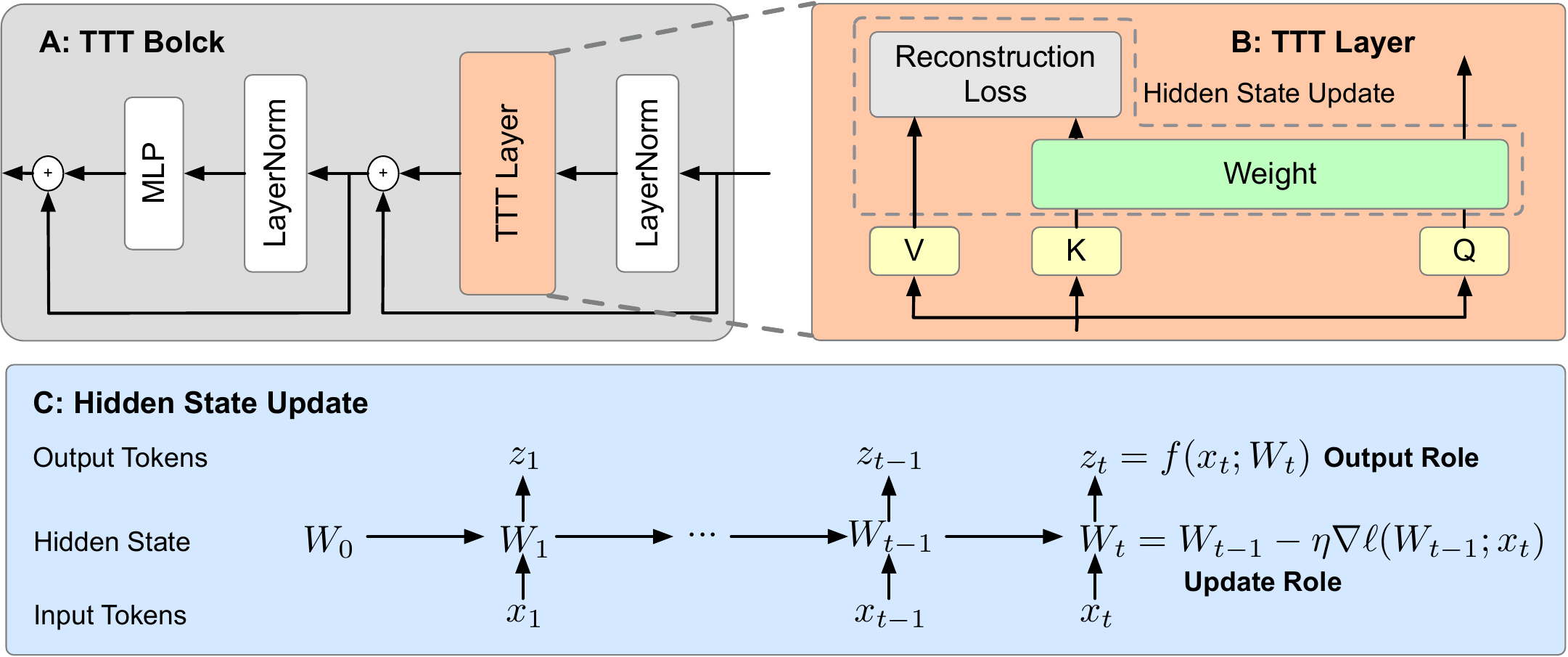} 
    \vspace{-1em}
    \caption{\textbf{A}: Original TTT Block. The basic building block of Transformers, originally based on self-attention, is replaced with the TTT Layer in the Transformer backbone. \textbf{B}: Original TTT Layer, driven by self-supervised loss, updates the weights adaptively. \textbf{C}: Hidden state update rule. The key idea of TTT is to make the hidden state itself a model $f$ with weights $W$ , and the update rule a gradient step on the self-supervised loss $\ell$. Therefore, updating the hidden state on a test sequence is equivalent to training the model $f$ at test time~\cite{sun2024learning}.}
    \vspace{-1.5em}
    \label{fig:tttlayer} 
\end{figure*}

In addition, manual annotations for AUs are cumbersome and costly, as they require trained coders to label each frame individually. Common AU datasets, i.e., DISFA~\cite{mavadati2013disfa} and BP4D~\cite{zhang2014bp4d}, only contain a limited number of subjects (27 and 41 subjects respectively). Current AU detection methods are predominantly evaluated using within-domain validation, where both training and testing data are sourced from the same dataset. While these methods often demonstrate strong within-domain performance, such results can obscure overfitting as shown in Figure~\ref{fig:performance_gap}, as their ability to generalize to unseen domains remains insufficiently explored. Cross-domain evaluations frequently reveal significant performance degradation when models are applied to datasets with variations in camera settings, environmental conditions, and subject demographics~\cite{ertugrul2019cross, ertugrul2020crossing}.

Although some studies~\cite{ertugrul2019cross, ertugrul2020crossing, yuan2025auformer, yin2024fg} have made efforts to improve cross-domain generalization, their scope remains limited. Existing approaches to enhance generalization include identity-aware training on diverse datasets, as demonstrated by IdenNet~\cite{tu2019idennet}, and domain adaptation techniques~\cite{yin2021self}, but these methods typically require access to the target domain. DeepFN~\cite{hernandez2022deepfn} addresses this issue by normalizing facial expressions onto a common template, but significant challenges persist in achieving robust cross-domain performance.
AUFormer~\cite{yuan2025auformer} leverages large-scale pretraining~\cite{radford2021learning} and parameter-efficient transfer learning to improve cross-domain robustness but does not incorporate targeted strategies to address domain shifts. Similarly, FG-Net~\cite{yin2024fg} utilizes generative model features to mitigate cross-domain challenges. Moreover, both methods rely on pretrained weights obtained from extensive additional data.

Apart from recent architectures such as Transformer and Mamba~\cite{gu2023mamba,xie2024fusionmamba}, the concept of learning at test time has a long history in machine learning~\cite{bottou1992local, gammerman2013learning}. More recently, Test-Time Training (TTT) extended this idea to Recurrent Neural Networks (RNNs)\cite{schuster1997bidirectional} and has emerged as an effective strategy for addressing out-of-distribution (OOD) problem\cite{gandelsman2022test, sun2024learning,meng2024scfusionttt}. The core idea of TTT is that each test instance defines its own learning problem, where the test instance itself serves as the target for generalization.
In traditional settings, a predictor $f$ is trained on all training instances and used to directly predict  $f(x)$ for a given test instance $x$ . In contrast, TTT formulates a specific learning problem centered on $x$. It updates a model $f_x$ using $x$ (with $f$ as its initialization) and then predicts $f_x(x)$. Since the test instance lacks labels, the learning problem is defined using a self-supervised task. Previous studies demonstrated that TTT with reconstruction tasks significantly enhances performance, particularly for outlier instances~\cite{gandelsman2022test}. 

In this paper, we introduce AU-TTT, a novel facial action unit detection method designed to enhance cross-domain generalization. AU-TTT leverages Test-Time Training to improve the model’s generalization ability. Additionally, the integration of bidirectional scanning and AU RoI scanning enables AU-TTT to effectively capture both general image information and fine-grained AU-related facial features, further enhancing its robustness and performance.

Our main contributions can be summarized as follows:
\begin{itemize}
\item We propose AU-TTT, a novel framework that integrates Test-Time Training into facial AU detection to address the challenge of out-of-distribution generalization.
\item We optimize the mini-batch strategy of the original TTT and propose a bidirectional scanning mechanism for image tasks.
\item We design AU RoI scanning mechanisms within AU-TTT tailored to the characteristics of AUs.
\item Our method demonstrates competitive performance in both within-domain and cross-domain scenarios.
\end{itemize}

\section{AU-TTT}

In this section, we first outline the fundamentals of the TTT Layer~\cite{sun2024learning}. We then present AU-TTT, our proposed adaptation of the Test-Time Training framework, specifically designed for facial AU detection.


\begin{figure*}[t]
    \centering
    \includegraphics[width=0.6\linewidth]{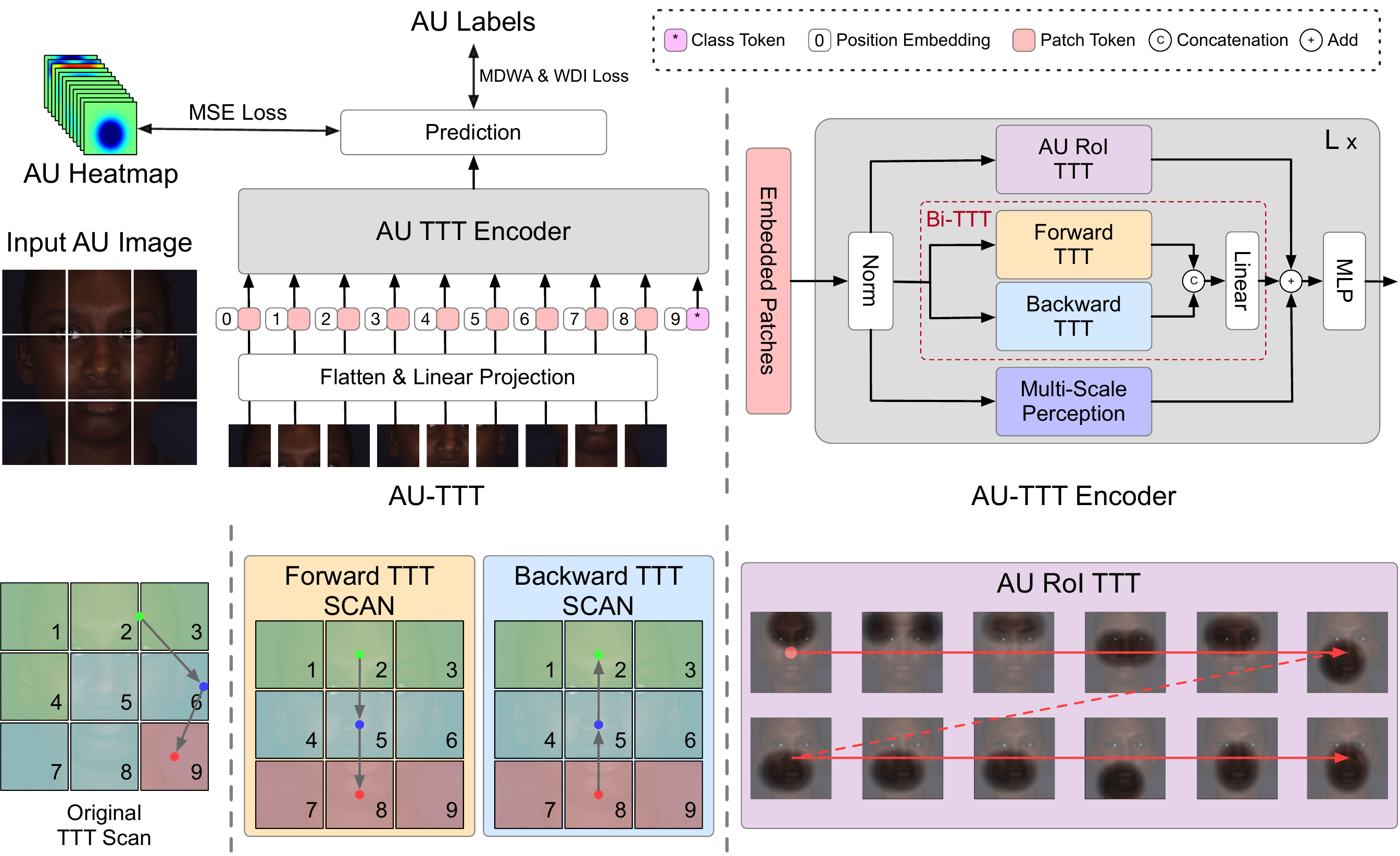} 
    \vspace{-1em}
    \caption{\textbf{Top Left}: The overall AU-TTT framework. \textbf{Top Right}: The architecture of the AU-TTT Block. \textbf{Bottom Left}: The original TTT Layer scanning method. \textbf{Bottom Center}: The bidirectional TTT scanning method. \textbf{Bottom Right}: The AU RoI TTT scanning method.}
    \vspace{-1.5em}
    \label{fig:framework} 
\end{figure*}

\subsection{Preliminaries}

We briefly introduce the original TTT Block and TTT Layer proposed in \cite{sun2024learning}, as illustrated in Figure~\ref{fig:tttlayer}. The TTT Block replaces the self-attention in the basic building block of Transformers with the TTT Layer. The TTT Layer leverages self-supervised reconstruction to encode information into the model’s weights.
This process is demonstrated using input sequences  $X = [x_t], t \in \{1, \dots, N\}$. It begins with an initial hidden state $W_0$, which is iteratively refined through self-supervised learning. For each input token  $x_t$, the model computes an output as follows
\begin{equation}
 z_t = f(x_t; W_t),
\end{equation}
where $W_t$ represents the current hidden state, and $z_t$ is the updated embedding. The hidden state $W_t$ is refined using a gradient descent step on a self-supervised loss $\ell$, which minimizes the difference between the model’s predicted embedding and the actual embedding. The update rule, illustrated in Figure~\ref{fig:tttlayer} (C), can also be expressed as
\begin{equation}
 W_t = W_{t-1} - \eta \nabla \ell(W_{t-1}; x_t),
\end{equation} 
where $\eta$ denotes the learning rate.
The TTT Layer further incorporates a multi-view reconstruction mechanism to enhance the optimization of the reconstruction process. The updated self-supervised loss is described as
\begin{equation}
\ell(W; x_t) = \left\| f(\theta_K x_t; W) - \theta_V x_t \right\|^2,
\end{equation} 
where \(\theta_K\) and \(\theta_V\) are the projections for the training view and label view, respectively. The corresponding output rule is as follows
\begin{equation}
z_t = f\left(\theta_Q x_t; W_t\right),
\end{equation} 
where \(\theta_Q\) is the projection for the test view. 

\subsection{AU-TTT}

The original TTT Layer was designed for NLP tasks and is not directly applicable to image-based AU detection. To address this limitation, we made several improvements in this subsection. The first issue to consider is that the mini-batch size within the original TTT Layer is suboptimal for CV tasks. Through experimentation, we discovered that adjusting the mini-batch size based on the number of tokens corresponding to the image’s feature rows significantly improves performance. We refer to this modified version of TTT as forward TTT. Next, we incorporate bidirectional scan TTT (Bi-TTT) into the AU-TTT block to mitigate the impact of sequential modeling on image tasks. Additionally, we introduce an AU RoI scanning mechanism (AU RoI TTT) specifically designed for AU detection tasks.

An overview of the proposed AU-TTT framework is shown in Figure~\ref{fig:framework}. The standard TTT Layer is originally designed for 1D sequences. To adapt it for vision tasks, we first transform the input image $\mathbf{I} \in \mathbb{R}^{H \times W \times C}$ into flattened patches $\mathbf{X} \in \mathbb{R}^{J \times (P^2 \cdot C)}$, where  $(H, W)$ denotes the height and width of the image, $C$ represents the number of channels, and $P$ is the size of each image patch. Next, we linearly project $\mathbf{X}$ into a vector of size $D$ and add learnable positional embeddings $\mathbf{E}_{\text{pos}} \in \mathbb{R}^{(J+1) \times D}$, as follows
\begin{equation}
  \mathbf{Z}_0 = \left[X^1 \mathbf{W}; X^2 \mathbf{W}; \cdots; X^J \mathbf{W};\text{CLS} \right] + \mathbf{E}_{\text{pos}},  
\end{equation}
where $X^j$ represents the $j$-th patch of $\mathbf{X}$, $\mathbf{W} \in \mathbb{R}^{(P^2 \cdot C) \times D}$ is the learnable projection. Inspired by ViT \cite{dosovitskiy2020image}, we also incorporate a class token (\text{CLS}) to represent the entire patch sequence. The token sequence $\mathbf{Z}_{l-1}$ is then passed through the  $l$-th layer of the AU-TTT encoder, producing the output $\mathbf{Z}_{l}$.

Specifically, the input token sequence $Z_{l - 1}$ is first normalized using a normalization layer. Then, the token sequence is processed in both forward and backward directions. The bidirectional TTT process is formulated as follows (for simplicity, the layer $l$ notation is omitted in the equations below):
\begin{equation}
\begin{aligned}
\text{Bi-TTT}(\mathbf{Z}) 
&= \text{Bi-TTT}([\mathbf{z}^1; \mathbf{z}^2; \dots; \mathbf{z}^J; \text{CLS}]) \\
&= \text{Linear}(\text{Cat}[\text{TTT}([\mathbf{z}^1; \mathbf{z}^2; \dots; \mathbf{z}^J; \text{CLS}]); \\
&\qquad \text{Reverse}(\text{TTT}([\mathbf{z}^J; \mathbf{z}^{J-1}; \dots; \mathbf{z}^1; \text{CLS}]))]).
\end{aligned}
\end{equation}
where $\mathbf{Z}$ represents the input sequence, $\mathbf{z}^j$ denotes the $j$-th token, and $\text{CLS}$ is the class token. The operation $\text{Reverse}(\cdot)$  reverses the sequence order, excluding the $\text{CLS}$ token. $\text{TTT}$ represents the foward TTT Layer, $\text{Cat}[\cdot;\cdot]$ denotes the concatenation of results along the channel dimension, and $\text{Linear}$ is a linear projection used to reduce the dimension.

Additionally, we leverage facial landmarks to generate a heatmap for extracting local features of AU RoI regions~\cite{yin2024fg}. Specifically, we first perform an element-wise multiplication between the facial features and the generated mask:
\begin{equation}
\mathbf{Z}_{\text{AU}} = \mathbf{Z} \cdot \mathbf{Mask},
\end{equation}
where $\mathbf{Z} \in \mathbb{R}^{H' \times W' \times D}$ and $\mathbf{Mask} \in \mathbb{R}^{\text{N}_\text{AU} \times H' \times W'}$. Here, $H'$ and $W'$ represent the height and width of the features, respectively. $\text{N}_\text{AU}$ represents the number of AUs in the dataset. Through broadcasting, the element-wise multiplication results in $\mathbf{Z}_{\text{AU}} \in \mathbb{R}^{\text{N}_\text{AU} \times H' \times W' \times D}$. Next, we compress $\mathbf{Z}_{\text{AU}}$ into tokens $\mathbf{Z}'_{\text{AU}} \in \mathbb{R}^{\text{N}_\text{AU} \times D}$ using mean pooling, which are then modeled using the TTT layer to capture the relationships between different AUs. 

At the same time, we employ a set of dilated convolutions with varying dilation rates as a Multi-Scale Perception (MSP) module to extract multi-scale facial features. Finally, the outputs of the AU RoI TTT, bidirectional TTT, and MSP modules are combined through addition and passed through a MLP to produce the output token sequence $\mathbf{Z}_l$.

\subsection{Loss Function}
AU detection is typically formulated as a multi-label binary classification problem. To generate the final predictions, we transform the $\text{CLS}$ token and the heatmap obtained from the last block of AU-TTT.
For the classification loss, we adopt the Margin-Truncated Difficulty-Aware Weighted Asymmetric Loss (MDWA-Loss) proposed by AUFormer~\cite{yuan2025auformer}. This loss function emphasizes activated AUs, adaptively evaluates the difficulty of unactivated AUs, and discards potentially mislabeled samples. The MDWA-Loss can be expressed as:
\begin{equation}
\begin{aligned}
\mathcal{L}_{\text{MDWA}} = -\frac{1}{N} \sum_{i=1}^N \omega_i & [ y_i \log(p_i) \\
& + (1 - y_i) p_{m,i}^{\gamma_i} \log(1 - p_{m,i}) ],
\end{aligned}
\label{equation:MDWAloss}
\end{equation}
where $y_i$ and $p_i$ represent the ground truth and predictions for the i-th AU, respectively. $\omega_i$ is the weight for addressing class imbalance, and it can be formulated as $\omega_i = N (1 / r_i) / {\sum_{j=1}^N (1 / r_j)}$, where $r_i$ is the occurrence rate of the i-th AU. $p_{m,i}$ is used to discard potentially mislabeled samples and can be formulated as $p_{m,i} = \max(p_i - m, 0)$, where $m \in [0, 1]$ is the truncation margin. $\gamma_i$ is used to distinguish the different difficulty levels of unactivated AUs and can be formulated as $\gamma_i = B_L + (B_R - B_L) \times r_i$, where $B_L$ and $B_R$ are left and right boundaries of $\gamma_i$ respectively.

Additionally, referring to \cite{shao2021jaa}, we also introduce a weighted multi-label dice loss (WDI-Loss) to alleviate the issue of AU prediction bias towards non-occurrence, as follow:
\begin{equation}
\mathcal{L}_{\text{WDI}} = \frac{1}{N} \sum_{i=1}^{N} \omega_i \left(1 - \frac{2 y_i p_i + \epsilon}{y_i^2 + p_i^2 + \epsilon} \right),
\label{equation:WDIloss}
\end{equation}
where $\epsilon$ is a smooth term and the meanings of $y_i$, $p_i$, and $\omega_i$ remain consistent with those defined in Equation~(\ref{equation:MDWAloss}).

\begin{table}[t] 
\centering
\caption{Average F1-score (in \%) results for within-domain evaluations on BP4D and DISFA. The best and second-best results for each column are bolded and underlined, respectively.}
\vspace{-1em}
\label{tab:DISFA_BP4D}

\begin{tabular}{l|c|cc}
\toprule
Method & Venue & BP4D & DISFA \\
\midrule
DRML~\cite{zhao2016deep} & CVPR 16 & 48.3 & 26.7 \\
LP-Net~\cite{niu2019local} & CVPR 19 & 61.0 & 56.9 \\
SRERL~\cite{li2019semantic} & AAAI 19 & 62.9 & 55.9 \\
J{\^A}A-Net~\cite{shao2021jaa} & IJCV 21 & 62.4 & 63.5 \\
HMP-PS~\cite{song2021hybrid} & CVPR 21 & 63.4 & 61.0 \\
SEV-Net~\cite{yang2021exploiting} & CVPR 21 & 63.9 & 58.8 \\
FAUDT~\cite{jacob2021facial} & CVPR 21 & 64.2 & 61.5 \\
PIAP~\cite{tang2021piap} & ICCV 21 & 64.1 & 63.8 \\
KDSRL~\cite{chang2022knowledge} & CVPR 22 & 64.5 & 64.5 \\
ME-GraphAU~\cite{luo2022learning} & IJCAI 22 & 65.5 & 62.4 \\ 
BG-AU~\cite{cui2023biomechanics} & CVPR 23 & 64.1 & 58.2 \\
KS~\cite{li2023knowledge} & ICCV 23 & 64.7 & 62.8 \\
FG-Net~\cite{yin2024fg} & WACV 24 & 64.3 & 65.4 \\ 
AUFormer~\cite{yuan2025auformer} & ECCV 24 & \textbf{66.2} & \textbf{66.4} \\
\midrule
\rowcolor{gray!30}
\textbf{AU-TTT (Ours)} & - & \underline{65.6} & \textbf{66.4} \\
\bottomrule
\end{tabular}
\vspace{-2em}
\end{table}

In addition, we utilize Mean Squared Error (MSE) loss to enhance the network’s capability in capturing AU RoI location-related information which can be expressed as:
\begin{equation}
\mathcal{L}_{MSE} = \| m - \hat{m} \|_2^2,
\end{equation}
where m represents the ground-truth heatmap and $\hat{m}$ represents the predicted heatmap. 

The overall loss of AUFormer can be expressed as:
\begin{equation}
\mathcal{L} = \lambda_{\text{MDWA}}\cdot\mathcal{L}_{\text{MDWA}} + \lambda_{\text{WDI}}\cdot\mathcal{L}_{\text{WDI}} + \lambda_{\text{MSE}}\cdot\mathcal{L}_{\text{MSE}},
\end{equation}
where $\lambda_{\text{MDWA}}$, $\lambda_{\text{WDI}}$, and $\lambda_{\text{MSE}}$ are hyperparameters.

\section{Experiments}

\subsection{Settings}

\textbf{Datasets.} We select two publicly available datasets, DISFA~\cite{mavadati2013disfa} and BP4D~\cite{zhang2014bp4d}, which are commonly used for AU detection. These datasets were collected from different subjects under varying backgrounds and lighting conditions. Follow~\cite{luo2022learning, shao2021jaa, he2016deep, yin2024fg, yuan2025auformer}, we employ subject-exclusive 3-fold cross-validation.

\begin{table*}[t]
\caption{F1-scores (in \%) for cross-domain evaluations between BP4D and DISFA. The best and second best results for each column are bolded and underlined, respectively. $^*$The numbers are derived from a replication of \cite{yin2024fg} based on open-source code.
}
\vspace{-1em}
\centering
\begin{tabular}{l|ccccc|c|ccccc|c}
\toprule
Direction & \multicolumn{6}{c|}{BP4D $\rightarrow$ DISFA} & \multicolumn{6}{c}{DISFA $\rightarrow$ BP4D} \\
\midrule
AU & 1 & 2 & 4 & 6 & 12 & \textbf{Avg.} & 1 & 2 & 4 & 6 & 12 & \textbf{Avg.} \\
\midrule
DRML$^*$ \cite{zhao2016deep} & 10.4 & 7.0 & 16.9 & 14.4 & 22.0 & 14.1 & 19.4 & 16.9 & 22.4 & 58.0 & 64.5 & 36.3  \\
J{\^A}A-Net$^*$ \cite{shao2021jaa} & 12.5 & 13.2 & 27.6 & 19.2 & 46.7 & 23.8 & 10.9 & 6.7 & 42.4 & 52.9 & 68.3 & 36.2  \\
ME-GraphAU$^*$ \cite{luo2022learning} & 43.3 & 22.5 & 41.7 & 23.0 & 34.9 & 33.1 & 36.5 & 30.3 & 35.8 & 48.8 & 62.2 & 42.7  \\
Patch-MCD \cite{yin2021self} & 34.3 & 16.6 & \underline{52.1} & 33.5 & 50.4 & 37.4 & - & - & - & - & - & -  \\
BG-AU~\cite{cui2023biomechanics} & - & - & - & - & - & 38.8 & - & - & - & - & - & -  \\
IdenNet \cite{tu2019idennet} & 20.1 & 25.5 & 37.3 & \textbf{49.6} & \textbf{66.1} & 39.7 & - & - & - & - & - & -  \\
FG-Net~\cite{yin2024fg}  & \textbf{61.3} & \textbf{70.5} & 36.3 & 42.2 & \underline{61.5} & \textbf{54.4} & \textbf{51.4} & \textbf{46.0} & 36.0 & 49.6 & 61.8 & 49.0 \\
AUFormer~\cite{yuan2025auformer} & \underline{49.7} & \underline{49.9} & 43.2 & 31.5 & 43.5 & 43.6 & \underline{49.8} & \underline{43.8} & \textbf{45.7} & \underline{66.6} & \underline{74.5} & \underline{56.1} \\
\midrule
\rowcolor{gray!30}
\textbf{AU-TTT (Ours)} & 46.0 & 44.6 & \textbf{56.7} & \underline{44.4} & 52.2 & \underline{48.7} & 49.4 & 39.5 & \underline{46.0} & \textbf{73.2} & \textbf{77.9} &  
\textbf{57.2} \\
\bottomrule
\end{tabular}
\vspace{-1.5em}
\label{tab:cross_domain}
\end{table*}

\textbf{Implementation Details.} We pretrain AU-TTT on ImageNet-1K~\cite{deng2009imagenet}, following the configuration of ViT-S/16~\cite{dosovitskiy2020image}. Specifically, we set the patch size to 16, the embedding dimension to 384, the depth to 12, and the number of heads to 6. The dilated rates $r_1$, $r_2$, and $r_3$ in Multi-Scale
Perception module are set to 1, 3, and 5. The left and right boundaries $B_L$ and $B_R$ of $\gamma$ are set to 1 and 2. The truncation margin $m$ for BP4D and DISFA are set to 0.1 and 0.15. The smooth term $\epsilon$ is set to 1. The number of AUs $\text{N}_\text{AU}$ for DISFA and BP4D is set to 8 and 12, following~\cite{zhao2016deep}.

\textbf{Evaluation Metric.} Following prior works~\cite{li2023knowledge, shao2021jaa, zhao2016deep, yin2024fg, yuan2025auformer}, we use the F1-score as the evaluation metric for AU detection performance. The F1-score represents the harmonic mean of precision and recall.

\subsection{Experimental Results}

The models are evaluated for both within-domain and cross-domain performance. Cross-domain evaluation is particularly critical for assessing the generalization capability of our AU detection approach.

\textbf{Within-domain Evaluation.}
We conduct within-domain evaluations on the widely used DISFA and BP4D datasets. Table~\ref{tab:DISFA_BP4D} reports the average F1-score results of various methods, demonstrating that AU-TTT achieves highly competitive performance on both BP4D and DISFA. Compared to previous CNN-based approaches and most Transformer-based methods, our approach not only delivers superior performance but also features linear complexity, effectively mitigating the quadratic complexity issue inherent to self-attention.

\textbf{Cross-domain Evaluation.}
To evaluate the generalization capabilities of AU-TTT, we conduct bidirectional cross-domain evaluations, specifically from BP4D to DISFA and from DISFA to BP4D. Following the approach of FG-Net~\cite{yin2024fg}, we use two folds of data from the source domain as the training set, while the entire dataset from the target domain serves as the testing set. As shown in Table~\ref{tab:cross_domain}, AU-TTT outperforms all methods in the DISFA-to-BP4D direction and ranks second only to FG-Net in the BP4D-to-DISFA direction, demonstrating highly competitive generalization capabilities.
It is noteworthy that FG-Net leverages features from StyleGAN2~\cite{karras2020analyzing}, pre-trained on FFHQ~\cite{karras2019style}, a large-scale dataset with high-quality and diverse facial data, which helps mitigate overfitting on BP4D. In contrast, our method relies only on pre-trained weights from ImageNet-1K, showcasing strong generalization with minimal dependence on additional data sources.

\subsection{Ablation Study}
We conduct ablation studies on the BP4D dataset without ImageNet-1K pretraining to avoid extensive pretraining costs and employ subject-exclusive 3-fold cross-validation to investigate the effects of each component of AU-TTT. Table~\ref{tab:AblationStudy} presents the average F1-score results for various combinations of AU-TTT’s components.

\textbf{Original TTT.}
We use a linear projection without the Test-Time Training strategy as a baseline. The first two rows of Table~\ref{tab:AblationStudy} demonstrate the effectiveness of the original TTT Layer.
\textbf{MSP.}
The second and third rows of Table~\ref{tab:AblationStudy} verify the importance of multi-scale perception (MSP), highlighting its role in leveraging multi-scale knowledge effectively.
\textbf{Foward TTT.} 
As shown in the third and fourth rows of Table~\ref{tab:AblationStudy}, adjusting the mini-batch size to better align with image tasks significantly enhances the performance of the TTT Layer on image tasks.
\textbf{Bidirectional TTT.} 
We ablate the bidirectional design of AU-TTT, as demonstrated in the fourth and fifth rows of Table~\ref{tab:AblationStudy}. The results indicate that the bidirectional strategy improves performance by  0.8\%  compared to the unidirectional AU-TTT.
\textbf{AU RoI TTT.} 
By comparing the last two rows of Table~\ref{tab:AblationStudy}, it is evident that incorporating additional AU RoI scanning enables the network to better extract AU-related local information, resulting in improved performance.

\begin{table}[t]
    \centering
    \caption{The average F1-score (in \%) results of various combinations of AU-TTT components on BP4D.}
    \label{tab:AblationStudy}
    \begin{tabular*}{\linewidth}{@{\extracolsep{\fill}}cccccc|c}
    \toprule[1pt]
    \multirow{2}{*}{Baseline} & Original & \multirow{2}{*}{MSP} & Foward & Backword  & AU RoI & \multirow{2}{*}{\textbf{Avg.}}  \\
     ~ & TTT & ~ & TTT & TTT & TTT & \\
    \midrule
    \cmark & ~ & ~ & ~ & ~ & ~ & 58.9 \\
    \cmark & \cmark & ~ & ~ & ~ & ~ & 60.1 \\
    \cmark & \cmark & \cmark & ~ & ~ & ~ & 60.4 \\
    \cmark & ~ & \cmark & \cmark & ~ & ~ & 61.2 \\
    \cmark & ~ & \cmark & \cmark & \cmark & ~ & 61.7 \\
    \cmark & ~ & \cmark & \cmark & \cmark & \cmark & 62.1 \\
    \bottomrule[1pt]
    \end{tabular*}
    \vspace{-1.5em}
\end{table}

\section{Conclusion}
In this paper, we propose AU-TTT, an effective and innovative method for generalizable AU detection. AU-TTT leverages Test-Time Training to enhance the model’s generalization ability and mitigate the overfitting caused by the limited availability of labeled AU datasets. By introducing TTT Layer and optimizing the scanning mechanism for images, AU-TTT effectively reduces the computational complexity associated with self-attention-based methods, particularly addressing the quadratic complexity challenge of Transformers. Additionally, the integration of an AU RoI scanning mechanism enables AU-TTT to better capture fine-grained, AU-related facial information, further improving its robustness and performance. These advancements make AU-TTT a promising approach for AU detection across diverse datasets and applications.




\bibliographystyle{IEEEbib}
\bibliography{references}

\begin{thebibliography}{10}

\bibitem{martinez2017automatic}
Brais Martinez, Michel~F Valstar, Bihan Jiang, and Maja Pantic,
\newblock ``Automatic analysis of facial actions: A survey,''
\newblock {\em IEEE transactions on affective computing}, vol. 10, no. 3, pp. 325--347, 2017.

\bibitem{xing2024emo}
Bohao Xing, Zitong Yu, et~al.,
\newblock ``Emo-llama: Enhancing facial emotion understanding with instruction tuning,''
\newblock {\em arXiv preprint arXiv:2408.11424}, 2024.

\bibitem{li2024enhancing}
Deng Li, Bohao Xing, and Xin Liu,
\newblock ``Enhancing micro gesture recognition for emotion understanding via context-aware visual-text contrastive learning,''
\newblock {\em IEEE Signal Processing Letters}, 2024.

\bibitem{li2024counterfactual}
Yong Li, Menglin Liu, et~al.,
\newblock ``Counterfactual discriminative micro-expression recognition,''
\newblock {\em Visual Intelligence}, vol. 2, no. 1, pp. 29, 2024.

\bibitem{li2024eald}
Deng Li, Xin Liu, Bohao Xing, Baiqiang Xia, Yuan Zong, Bihan Wen, and Heikki K{\"a}lvi{\"a}inen,
\newblock ``Eald-mllm: Emotion analysis in long-sequential and de-identity videos with multi-modal large language model,''
\newblock {\em arXiv preprint arXiv:2405.00574}, 2024.

\bibitem{he2016deep}
Kaiming He, Xiangyu Zhang, et~al.,
\newblock ``Deep residual learning for image recognition,''
\newblock in {\em Proceedings of the IEEE/CVF Conference on CVPR}, 2016, pp. 770--778.

\bibitem{scarselli2008graph}
Franco Scarselli, Marco Gori, et~al.,
\newblock ``The graph neural network model,''
\newblock {\em IEEE Transactions on Neural Networks}, vol. 20, no. 1, pp. 61--80, 2008.

\bibitem{jacob2021facial}
Geethu~Miriam Jacob and Bjorn Stenger,
\newblock ``Facial action unit detection with transformers,''
\newblock in {\em Proceedings of the IEEE/CVF Conference on CVPR}, 2021, pp. 7680--7689.

\bibitem{liu2024multi}
Xin Liu, Kaishen Yuan, et~al.,
\newblock ``Multi-scale promoted self-adjusting correlation learning for facial action unit detection,''
\newblock {\em IEEE Transactions on Affective Computing}, 2024.

\bibitem{luo2022learning}
Cheng Luo, Siyang Song, et~al.,
\newblock ``Learning multi-dimensional edge feature-based au relation graph for facial action unit recognition,''
\newblock {\em arXiv preprint arXiv:2205.01782}, 2022.

\bibitem{wang2022semantic}
Xuehan Wang, CL~Philip Chen, et~al.,
\newblock ``Semantic learning for facial action unit detection,''
\newblock {\em IEEE Transactions on Computational Social Systems}, vol. 10, no. 3, pp. 1372--1380, 2022.

\bibitem{yang2023fan}
Jing Yang, Jie Shen, et~al.,
\newblock ``Fan-trans: Online knowledge distillation for facial action unit detection,''
\newblock in {\em Proceedings of the IEEE/CVF WACV}, 2023, pp. 6019--6027.

\bibitem{yuan2025auformer}
Kaishen Yuan, Zitong Yu, et~al.,
\newblock ``Auformer: Vision transformers are parameter-efficient facial action unit detectors,''
\newblock in {\em Proceedings of the European Conference on Computer Vision}. Springer, 2025, pp. 427--445.

\bibitem{wang2024multi}
Zihan Wang, Siyang Song, et~al.,
\newblock ``Multi-scale dynamic and hierarchical relationship modeling for facial action units recognition,''
\newblock in {\em Proceedings of the IEEE/CVF Conference on CVPR}, 2024, pp. 1270--1280.

\bibitem{dosovitskiy2020image}
Alexey Dosovitskiy,
\newblock ``An image is worth 16x16 words: Transformers for image recognition at scale,''
\newblock {\em arXiv preprint arXiv:2010.11929}, 2020.

\bibitem{zhao2016deep}
Kaili Zhao, Wen-Sheng Chu, et~al.,
\newblock ``Deep region and multi-label learning for facial action unit detection,''
\newblock in {\em Proceedings of the IEEE/CVF Conference on CVPR}, 2016, pp. 3391--3399.

\bibitem{shao2021jaa}
Zhiwen Shao, Zhilei Liu, et~al.,
\newblock ``Jaa-net: joint facial action unit detection and face alignment via adaptive attention,''
\newblock {\em IJCV}, vol. 129, pp. 321--340, 2021.

\bibitem{yin2024fg}
Yufeng Yin, Di~Chang, et~al.,
\newblock ``Fg-net: Facial action unit detection with generalizable pyramidal features,''
\newblock in {\em Proceedings of the IEEE/CVF WACV}, 2024, pp. 6099--6108.

\bibitem{niu2019local}
Xuesong Niu, Hu~Han, et~al.,
\newblock ``Local relationship learning with person-specific shape regularization for facial action unit detection,''
\newblock in {\em Proceedings of the IEEE/CVF Conference on CVPR}, 2019, pp. 11917--11926.

\bibitem{bresson2017residual}
Xavier Bresson and Thomas Laurent,
\newblock ``Residual gated graph convnets,''
\newblock {\em arXiv preprint arXiv:1711.07553}, 2017.

\bibitem{sun2024learning}
Yu~Sun, Xinhao Li, et~al.,
\newblock ``Learning to (learn at test time): Rnns with expressive hidden states,''
\newblock {\em arXiv preprint arXiv:2407.04620}, 2024.

\bibitem{mavadati2013disfa}
S~Mohammad Mavadati, Mohammad~H Mahoor, et~al.,
\newblock ``Disfa: A spontaneous facial action intensity database,''
\newblock {\em IEEE Transactions on Affective Computing}, vol. 4, no. 2, pp. 151--160, 2013.

\bibitem{zhang2014bp4d}
Xing Zhang, Lijun Yin, et~al.,
\newblock ``Bp4d-spontaneous: a high-resolution spontaneous 3d dynamic facial expression database,''
\newblock {\em Image and Vision Computing}, vol. 32, no. 10, pp. 692--706, 2014.

\bibitem{ertugrul2019cross}
Itir~Onal Ertugrul, Jeffrey~F Cohn, et~al.,
\newblock ``Cross-domain au detection: Domains, learning approaches, and measures,''
\newblock in {\em Proceedings of the IEEE International Conference on Automatic Face \& Gesture Recognition (FG)}. IEEE, 2019, pp. 1--8.

\bibitem{ertugrul2020crossing}
Itir~Onal Ertugrul, Jeffrey~F Cohn, et~al.,
\newblock ``Crossing domains for au coding: Perspectives, approaches, and measures,''
\newblock {\em IEEE Transactions on Biometrics, Behavior, and Identity Science}, vol. 2, no. 2, pp. 158--171, 2020.

\bibitem{tu2019idennet}
Cheng-Hao Tu, Chih-Yuan Yang, et~al.,
\newblock ``Idennet: Identity-aware facial action unit detection,''
\newblock in {\em Proceedings of the IEEE International Conference on Automatic Face \& Gesture Recognition (FG)}. IEEE, 2019, pp. 1--8.

\bibitem{yin2021self}
Yufeng Yin, Liupei Lu, et~al.,
\newblock ``Self-supervised patch localization for cross-domain facial action unit detection,''
\newblock in {\em Proceedings of the IEEE International Conference on Automatic Face \& Gesture Recognition (FG)}. IEEE, 2021, pp. 1--8.

\bibitem{hernandez2022deepfn}
Javier Hernandez, Daniel McDuff, et~al.,
\newblock ``Deepfn: towards generalizable facial action unit recognition with deep face normalization,''
\newblock in {\em Proceedings of the International Conference on Affective Computing and Intelligent Interaction (ACII)}. IEEE, 2022, pp. 1--8.

\bibitem{radford2021learning}
Alec Radford, Jong~Wook Kim, et~al.,
\newblock ``Learning transferable visual models from natural language supervision,''
\newblock in {\em Proceedings of the International Conference on Machine Learning}. PMLR, 2021, pp. 8748--8763.

\bibitem{gu2023mamba}
Albert Gu and Tri Dao,
\newblock ``Mamba: Linear-time sequence modeling with selective state spaces,''
\newblock {\em arXiv preprint arXiv:2312.00752}, 2023.

\bibitem{xie2024fusionmamba}
Xinyu Xie, Yawen Cui, et~al.,
\newblock ``Fusionmamba: Dynamic feature enhancement for multimodal image fusion with mamba,''
\newblock {\em Visual Intelligence}, vol. 2, no. 1, pp. 37, 2024.

\bibitem{bottou1992local}
L{\'e}on Bottou and Vladimir Vapnik,
\newblock ``Local learning algorithms,''
\newblock {\em Neural computation}, vol. 4, no. 6, pp. 888--900, 1992.

\bibitem{gammerman2013learning}
Alex Gammerman, Volodya Vovk, et~al.,
\newblock ``Learning by transduction,''
\newblock {\em arXiv preprint arXiv:1301.7375}, 2013.

\bibitem{schuster1997bidirectional}
Mike Schuster and Kuldip~K Paliwal,
\newblock ``Bidirectional recurrent neural networks,''
\newblock {\em IEEE Transactions on Signal Processing}, vol. 45, no. 11, pp. 2673--2681, 1997.

\bibitem{gandelsman2022test}
Yossi Gandelsman, Yu~Sun, et~al.,
\newblock ``Test-time training with masked autoencoders,''
\newblock {\em Advances in NeurIPS}, vol. 35, pp. 29374--29385, 2022.

\bibitem{meng2024scfusionttt}
Dian Meng, Bohao Xing, Xinlei Huang, Yanran Liu, Yijun Zhou, Zitong Yu, Xubin Zheng, et~al.,
\newblock ``scfusionttt: Single-cell transcriptomics and proteomics fusion with test-time training layers,''
\newblock {\em arXiv preprint arXiv:2410.13257}, 2024.

\bibitem{li2019semantic}
Guanbin Li, Xin Zhu, et~al.,
\newblock ``Semantic relationships guided representation learning for facial action unit recognition,''
\newblock in {\em Proceedings of the AAAI Conference on Artificial Intelligence}, 2019, vol.~33, pp. 8594--8601.

\bibitem{song2021hybrid}
Tengfei Song, Zijun Cui, et~al.,
\newblock ``Hybrid message passing with performance-driven structures for facial action unit detection,''
\newblock in {\em Proceedings of the IEEE/CVF Conference on CVPR}, 2021, pp. 6267--6276.

\bibitem{yang2021exploiting}
Huiyuan Yang, Lijun Yin, Yi~Zhou, et~al.,
\newblock ``Exploiting semantic embedding and visual feature for facial action unit detection,''
\newblock in {\em Proceedings of the IEEE/CVF Conference on CVPR}, 2021, pp. 10482--10491.

\bibitem{tang2021piap}
Yang Tang, Wangding Zeng, et~al.,
\newblock ``Piap-df: Pixel-interested and anti person-specific facial action unit detection net with discrete feedback learning,''
\newblock in {\em Proceedings of the IEEE/CVF ICCV}, 2021, pp. 12899--12908.

\bibitem{chang2022knowledge}
Yanan Chang and Shangfei Wang,
\newblock ``Knowledge-driven self-supervised representation learning for facial action unit recognition,''
\newblock in {\em Proceedings of the IEEE/CVF Conference on CVPR}, 2022, pp. 20417--20426.

\bibitem{cui2023biomechanics}
Zijun Cui, Chenyi Kuang, et~al.,
\newblock ``Biomechanics-guided facial action unit detection through force modeling,''
\newblock in {\em Proceedings of the IEEE/CVF Conference on CVPR}, 2023, pp. 8694--8703.

\bibitem{li2023knowledge}
Xiaotian Li, Xiang Zhang, et~al.,
\newblock ``Knowledge-spreader: Learning semi-supervised facial action dynamics by consistifying knowledge granularity,''
\newblock in {\em Proceedings of the IEEE/CVF ICCV}, 2023, pp. 20979--20989.

\bibitem{deng2009imagenet}
Jia Deng, Wei Dong, et~al.,
\newblock ``Imagenet: A large-scale hierarchical image database,''
\newblock in {\em Proceedings of the IEEE/CVF Conference on CVPR}. Ieee, 2009, pp. 248--255.

\bibitem{karras2020analyzing}
Tero Karras, Samuli Laine, et~al.,
\newblock ``Analyzing and improving the image quality of stylegan,''
\newblock in {\em Proceedings of the IEEE/CVF Conference on CVPR}, 2020, pp. 8110--8119.

\bibitem{karras2019style}
Tero Karras,
\newblock ``A style-based generator architecture for generative adversarial networks,''
\newblock {\em arXiv preprint arXiv:1812.04948}, 2019.

\end{thebibliography}

\end{document}